\documentclass[letterpaper, 10 pt, conference]{ieeeconf}  
\IEEEoverridecommandlockouts
\usepackage{cite}
\usepackage{amsmath,amssymb,amsfonts}
\usepackage{algorithmic}
\usepackage{graphicx}
\usepackage{textcomp}
\usepackage{array}
\usepackage{float}
\usepackage{flushend}
\def\BibTeX{{\rm B\kern-.05em{\sc i\kern-.025em b}\kern-.08em
    T\kern-.1667em\lower.7ex\hbox{E}\kern-.125emX}}
\begin{document}

\title{DFNet: Semantic Segmentation on Panoramic Images with Dynamic Loss Weights and Residual Fusion Block\\
}

\author{Wei Jiang, Yan Wu$^{*}$
\thanks{This work is supported by the National Key Research and Development Project (No.2017YFC0821300)}%
\thanks{All authors are with the Department of Computer Science and Technology, School of Electronics and Information Engineering, Tongji University, Shanghai, 201804, China
        {\tt\small yanwu@tongji.edu.cn}}}

\maketitle

\begin{abstract}
For the self-driving and automatic parking, perception is the basic and critical technique, moreover, the detection of lane markings and parking slots is an important part of visual perception. 
In this paper, we use the semantic segmentation method to segment the area and classify the class of lane makings and parking slots on panoramic surround view (PSV) dataset. 
We propose the DFNet and make two main contributions, one is dynamic loss weights, and the other is residual fusion block (RFB).
Dynamic loss weights are varying from classes, calculated according to the pixel number of each class in a batch. 
RFB is composed of two convolutional layers, one pooling layer, and a fusion layer to combine the feature maps by pixel multiplication. 
We evaluate our method on PSV dataset, and achieve an advanced result.
\end{abstract}


\section{INTRODUCTION}
In artificial intelligent field, automatic vehicle is being studied in great demand both in academic and industrial sector. 
In recent years, a large and fast-growing number of companies and colleges set up the team special for self-driving, and have launched their own automatic vehicles. 
It is significant for self-driving to reduce the stress of drivers and improve the road safety. 
Self-driving technique can be divided into three parts, environmental perception, planning decision, execution control, and perception is the basis for autonomous systems. 
Environmental perception mainly includes visual perception and radar perception, while visual perception is currently more widely used.

As a branch of self-driving, automatic parking requires the correct information of lane makings and parking slots in the perception process. 
Comparing with front sight images, panoramic images can capture the surroundings of car completely; therefore it is more suitable to use panoramic images in a low speed environment like automatic parking.

In recent years, deep learning \cite{lecun2015deep} has made significant breakthroughs in image processing. 
For traditional visual algorithms, it needs to set the rules of extracting features based on prior knowledge, while the rules are complex and hard to be adaptive and robust. 
As for deep learning, it can automatically extract features from sample data with training, and got much better results as long as the sample data is sufficient enough. 
Visual perception in self-driving is essentially image processing, due to the high accuracy and strong robustness of deep learning, the technology of applying it to visual perception has become the current trend. 
A large number of traffic scene datasets \cite{6836060}\cite{cordts2016cityscapes} and corresponding methods of object detection \cite{ren2017faster}\cite{redmon2016you} and semantic segmentation \cite{badrinarayanan2017segnet}\cite{zhao2017pyramid} based on deep leaning are proposed, and proved to be useful.
Therefore, we use the semantic segmentation method of deep learning to segment the area and classify the class of lane makings and parking slots on panoramic images. 
To improve the accuracy of results, we propose the DFNet for semantic segmentation and make two main contributions:
\begin{enumerate}
	\item One is dynamic loss weights. 
	When computing the loss, we assign the weights to each class, which are calculated according to the pixel number of each class, to overcome the imbalance of different class.
	\item The other is residual fusion block (RFB). 
	It is used to refine the segmentation area, and reduce the classification error of pixels at the boundary between the two areas.
\end{enumerate}

\section{RELATED WORK}


In 2006, K Kato et al. \cite{kato2006image} firstly proposed a panoramic parking system, which can effectively eliminate visual blind area and then improve the efficiency and safety of parking. 
YC Liu et al. \cite{liu2008bird} presented a driving assistant system which can provide the bird’s eye view image of vehicle surroundings with six fisheye cameras. 
The multiple fisheye cameras were mounted around the vehicle to capture images of the surroundings. 
Because of the complete range of vision, various vision methods for lane markings and parking slots detection were proposed on panoramic images. 
C Wang \cite{wang2014automatic} extracted the parking slots with a Radon transform based method, JK Suhr and HG Jung \cite{suhr2014sensor} detected parking slots by exploiting a hierarchical tree structure of it and combining sequential detection results, HH Chi and LY Hsu \cite{hao2016dynamic} detected lane markings with line detection method. 
For the detection on panoramic images, traditional image processing methods are the main methods at present, the accuracy of which are closely related to the structure of the markings, and will be greatly influenced by the noise in the images, like a shade or a fuzzy structure.

Deep learning has achieved far more accurate results than traditional methods on image processing. 
There are several works on lane makings and parking slots detection using deep learning on front sight images.
J Kim and M Lee \cite{kim2014robust} presented a robust lane detection method based on the combined convolutional neural network with random sample consensus algorithm; S Lee et al.\cite{lee2017vpgnet} proposed a unified end-to-end trainable multi-task network that jointly handles lane and road marking detection and recognition guided by a vanishing point; G Amato et al. \cite{amato2017deep} proposed a decentralized and efficient solution for visual parking slots occupancy detection based on a deep convolutional neural network.

Traditional image processing methods on panoramic images and deep learning methods on front sight images both achieve excellent results, but few work is aimed to use deep leaning on panoramic images. 
This is because the accuracy of deep learning model is largely related to datasets, and there are few public dataset of panoramic images which can be used to train a model. 
The first public panoramic dataset for lane markings and parking slots is panoramic surround view (PSV) dataset, released by Yan Wu et al. \cite{wu2018vh}. 
And this dataset is specially used for semantic segmentation, which is labeled pixel by pixel, and each pixel has its corresponding class.
In order to combine the advantages of panoramic images and deep learning, we use the semantic segmentation method with convolutional network to segment the area and classify the class of lane makings and parking slots on the PSV dataset.

Semantic segmentation is the natural step to achieve fine-grained inference, its goal is to make dense predictions inferring labels for every pixel \cite{garcia2017review}. 
The most successful model for semantic segmentation is the fully convolutional network (FCN) by Long et al. \cite{long2015fully}, which is the first end-to-end semantic segmentation model, realized by enlarging the feature maps to the same size as input image. 
After that, all the network models are designed as end-to-end, which has the advantage that they do not need the post-processing. 
There are three main ways to enlarge the feature maps, deconvolution \cite{zeiler2014visualizing}, unpooling, and bilinear interpolation. 
The way used in FCN is deconvolution, which is the reverse process of convolution. 
In Segnet\cite{badrinarayanan2017segnet}, Enet\cite{paszke2016enet}, they used unpooling, which need positon parameters of pool mask from the corresponding pooling. 
To improve the accuracy, more complex models are proposed. With hundreds of layers, Resnet\cite{he2016deep} and Densenet\cite{jegou2017one} become the most common basic model used in convolutional networks. 
FCCN\cite{wu2017fully} added soft weights of cost function on different target objects; HFCN\cite{yang2018semantic} proposed a highly fused convolutional network with multiple soft cost functions;
Refinenet\cite{lin2017refinenet} trained the model with multiple scale of input images; F Yu \cite{yu2015multi} proposed dilation convolution to enlarge the receptive field of convolution without increase in parameters; PSPnet\cite{zhao2017pyramid} presented a pyramid pooling model; GCN\cite{peng2017large} applied large-size kernels and residual-based boundary refinement blocks.

In \cite{wu2018vh}, they achieved the segmentation of lane markings and parking slots using semantic segmentation method on PSV dataset, and proposed a VH-stage module special for linear structures. 
But the size of their model is too large to meet the requirement of using in embedded and mobile platform. 
In this paper, we put forward a smaller size model, but with higher accuracy. 
And the two main improved methods we proposed are proved to be significant.

\section{METHODS}

\subsection{network}\label{AA}
\begin{figure*}[ht]
\centering
\makebox[\textwidth][c]{\includegraphics[width=0.7\textwidth]{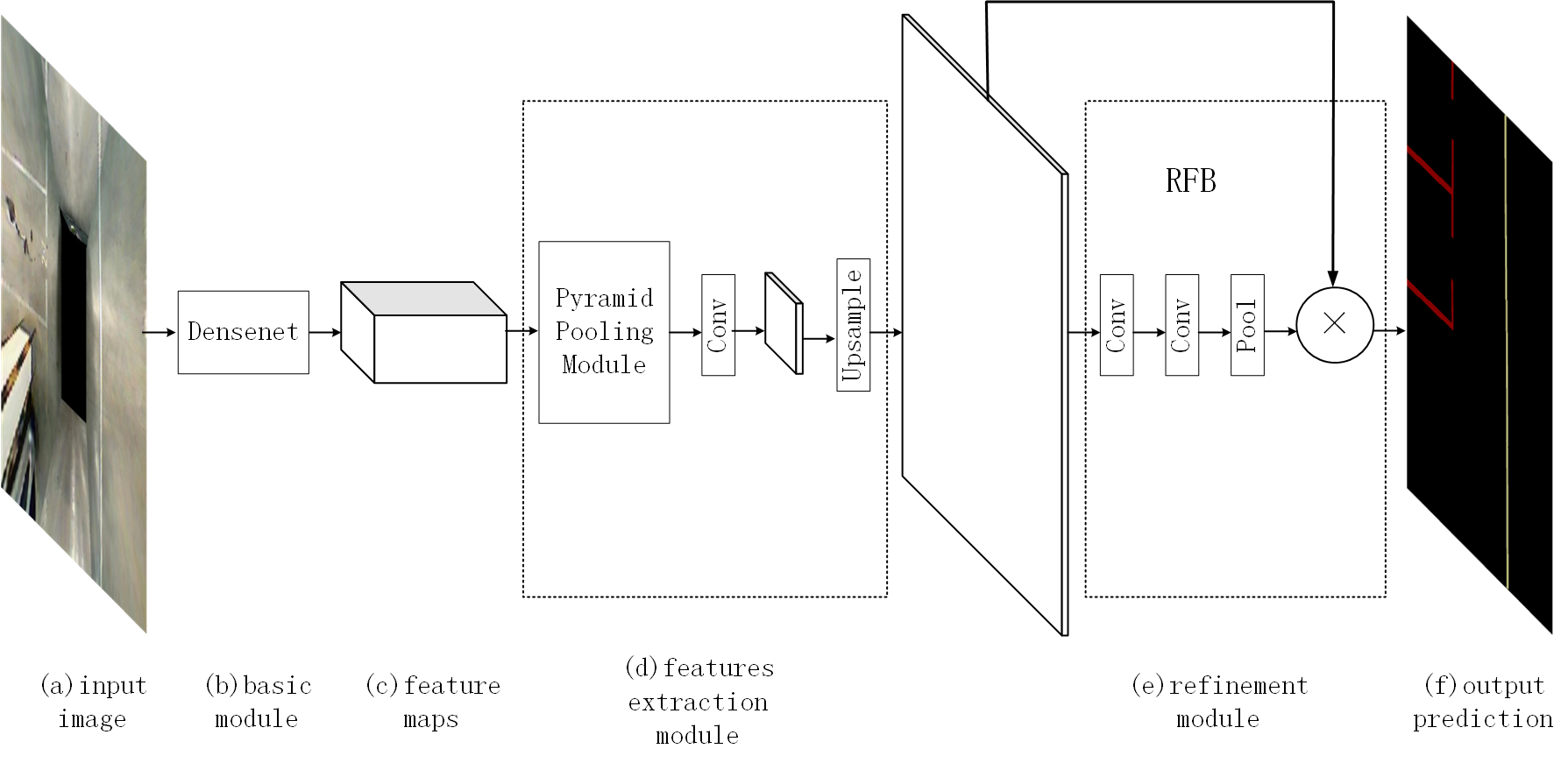}}
\caption{Overview of our proposed DFNet. Given an input image (a), we first use Densenet (b) to get the feature maps (c), then extract features by a pyramid pooling module and convolution layers(d), and enlarge the feature maps to the same size as input image. Finally, the RFB (e) is used to refine the segment area and then get the output (f).}
\label{network}
\end{figure*}

The proposed model, dense fusion network (DFNet), is illustrated in Fig.\ref{network}.
DFNet is adapted from PSPNet \cite{zhao2017pyramid}, which used to be the state-of-the-art model of semantic segmentation for a long time. 
DFNet can be divided into three parts, basic module, features extraction module, and refinement module. 
For basic module, we use the Densenet \cite{jegou2017one} as basic network. 
Comparing with Resnet \cite{he2016deep} used in many semantic segmentation models, Densenet has smaller model size and faster training speed, but similar accuracy. 
For features extraction module, we use the pyramid pooling module proposed by PSPNet, followed by convolutional layers and an upsample layer using bilinear interpolation. 
After these two modules, the feature maps are enlarged to the same size as input image. 
However, when the enlargement factor is large, it will bring noise and make the pixels at the boundary of two areas difficult to classify. 
Therefore, we add refinement module at the end of the model to refine the segmentation area. 
For refinement module, we propose a residual fusion block (RFB) which is the combination of convolution layers and pooling layers.

\begin{figure}[ht]
\centering
\includegraphics[height = 0.5\textwidth]{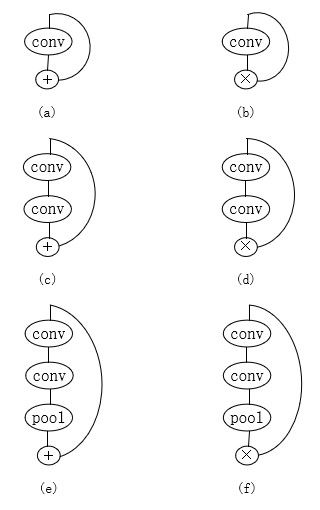}
\caption{There are several structures of RFB in our experiment. + and x represent the fusion ways of averaging and multiplying respectively.}
\label{RFB}
\end{figure}

RFB is used to refine the segmentation area of each class and reduce the influence of noise caused by enlarge layers. 
RFB is mainly focused on the classification of the pixels at the boundary between two areas, because these pixels are relatively difficult to classify, RFB can reduce the error prediction of these pixels, and then improve the accuracy. 
The main idea of RFB to divide the feature maps into two paths is similar to residual block. 
One path consists of convolutional layers or pooling layers, while the other has not any processing. 
Finally, we fuse the feature maps of these two paths by averaging or multiplying. 
This is because after processing with convolutional layers or pooling layers, the values of the points in feature maps will be slightly changed, while different degrees of change in different areas. 
The closer to the boundary, the greater the change is. 
By fusing the feature maps of two paths, the value of points with a greater difference will be corrected. 
We attempt several structures, all are displayed in Fig.\ref{RFB}. 
We will describe the configuration and effect in detail in C part of section IV.


\subsection{dynamic loss weights}
In the process of convolution neural network training, the value of network weights are adjusted by error calculated in loss function. 
But for the reason that the numbers of pixels in images vary from class to class, the influence of each class to the loss is different. 
The more the pixel number of a class is, the more the impact of this class on loss. 
For overcoming the imbalance of different class, we assign a weight to each class when calculate the loss. 
In Segnet \cite{badrinarayanan2017segnet}, they compute the weights according to the whole training set. 
But the network weights are adjusted after each iteration of training, and the pixel number of each class in a batch may be much different from that in the whole training set. 
So in each iteration, we calculate the weights according to the current input batch, and the weights are different in each iteration. 
The weights calculation formula is shown in Eq.\ref{equ1}.
\begin{equation}
w_i = \left\{ 
             \begin{array}{lc}  
             1, & n_i = 0	\\
             \beta,	&	w_i < \beta\\	
             \dfrac{N}{2*c*n_i},	&	\beta < w_i < \alpha\\
             \alpha,	&	w_i > \alpha
             \end{array}  
\right.  
\label{equ1}
\end{equation}

In the formula, $w_i$ is the weight of class $i$, $c$ is the class number, and the value of $i$ is from 0 to $c$.
$\beta$ and $\alpha$ are the lower and upper threshold of $w_i$, we set the threshold to avoid excessive weights differences. 
$N$ is the total pixel number of this batch, $n_i$ is the pixel number of class $i$, when $n_i$ = 0, it means that the class $i$ does not appear in this batch, we set the weight to 1. 
Because we need to increase the effect of small pixel number class on loss, so the smaller the $n_i$, the larger the $w_i$ is. $N$ and $c$ are constant, $w_i$ is just changed by $n_i$. 
When the $n_i$ is the average number, $w_i$ is calculated to be $\dfrac{1}{2}$, the multiplicative coefficient of $\dfrac{1}{2}$ is also used to decrease the $w_i$ of large pixel number of class.
The loss function is shown in Eq.\ref{equ2}, where $x_{ij}$, $y_{ij}$ are prediction class and label in pixel (i, j), $w$ is the loss weights.

\begin{equation}
LOSS = \sum_{i = 1}\sum_{j = 1}w\left \| x_{ij} - y_{ij} \right \|^2
\label{equ2}
\end{equation}

\section{EXPERINMENT}

\subsection{ experimental setup}
In our experiments, we evaluate our methods on PSV datasets \cite{wu2018vh}, which is made and released by The Tongji Intelligent Electric Vehicle (TiEV) team. 
The images are collected in Tongji university with two size, 600x600 and 1000x1000. 
There are a total of 4249 panoramic RGB images with labeled ground truth of 6 object classes, background, parking slots, white solid line, white dashed line, yellow solid line and yellow dashed line.
Thereinto, the number of images in train set, test set and valid set are 2550, 1274, 425. 
Our experiments are implemented on pytorch, and trained on NVIDIA GeForce GTX TITAN X graphic card. 
We crop the input images to a unified size of 600x600. 
Three metrics are used for evaluating our methods, pixel accuracy (pacc), mean pixel accuracy (mpacc), and mean intersection over union (mIoU).

The experiments are divided to three steps: firstly, we train our model with dynamic weights but without RFB, to get the optimum threshold in weights formula; secondly, using the threshold determined in the previous step, we verify these several RFBs to find the best structure; finally, we change the aux loss to further improve the accuracy.  

\subsection{ evaluation with dynamic loss weights}
There are two thresholds in weights formula, $\beta$ for the lower and $\alpha$ for the upper. 
Because only the pixel number of background is way above the average, so we assign the value of $\beta$ to 0.1, and change the value of $\alpha$ to find the best. 
We also train our model without dynamic loss weights as a reference. 
The results are shown in Table.\ref{tab1}.
From the Table.\ref{tab1}, we can see that, the dynamic loss weights indeed greatly improve the results. 
When the $\alpha$ is too large, the improvement is just a little. 
The excessive weight of one class results in too much difference among different class, which is equivalent to making a mutual transformation between the minority and the majority, instead of balancing the minority and the majority. 
When $\alpha$ is within 10, the improvements are more obvious. 
Thereinto, we get best result when $\alpha$ is 5, mIoU shows the improvements about 7.7\% comparing with not using dynamic weights.
Finally, we choose 5 as the value of $\alpha$.
\begin{table}[htbp]
\caption{The results with different loss weights}
\begin{center}
\begin{tabular}{cccc}
\hline
\textbf{$w$} & \textbf{pacc(\%)} & \textbf{mpacc(\%)} & \textbf{mIoU(\%)} \\
\hline
-,- & 97.67 & 41.77 & 36.06 \\
$\beta$ = 0.1, $\alpha$ = 3 & 95.99 & 81.19 & 42.86 \\
$\beta$ = 0.1, $\alpha$ = 5 & 95.73& 84.58& \textbf{43.79} \\
$\beta$ = 0.1, $\alpha$ = 7 & 95.32 & 86.10 & 43.31 \\
$\beta$ = 0.1, $\alpha$ = 10 & 95.99 & 81.19 & 42.86 \\
$\beta$ = 0.1, $\alpha$ = 50 & 93.64 & 88.91 & 38.87 \\
\hline
\end{tabular}
\label{tab1}
\end{center}
\end{table}

\subsection{ evaluation with refinement block}
There are six kinds of structures showed in Fig \ref{RFB}, the corresponding configuration and results are listed in Table \ref{tab2}, all the results are based on dynamic weights and $\alpha$ is 5 .
The difference of each structure lies in the combination of layers in one path and the fusion way of feature maps. 
Because the value of each point in the final feature maps is from 0 to 1, so after the fusion by averaging or multiplying, the value is also from 0 to 1.
From the table, we can see that, when using a single layer, the influence of the two fusion ways is not very different, but when using multiple layers, the multiplying way is better than averaging. 
In addition, when the number of layers is less or the fusion way is averaging, RFB has little improvements on the results. 
The best results is to use the structure of (f), which consists of two convolution layers and an average pooling layer, and it get 2.6\% improvements on mIoU comparing with not using RFB.
Finally we choose the last structure (f) as RFB.
\begin{table}[]
\caption{The configuration and results of different structures. k, d, p, s are kernel size, dilation, padding, stride respectively. The first row is the results without RFB. }
\begin{center}
\begin{tabular}{p{0.2cm}<{\centering}p{3.65cm}<{\centering}p{0.5cm}<{\centering}p{0.6cm}<{\centering}p{0.6cm}<{\centering}p{0.5cm}<{\centering}}
\hline
& \textbf{configuration} & \textbf{fusion} & \textbf{pacc} & \textbf{mpacc} & \textbf{mIoU} \\
&  &  & \textbf{(\%)} &  \textbf{ (\%)} & \textbf{(\%)} \\
\hline
&\\
- & --,-- & --,-- & 95.73 & 84.58 & 43.79 \\
&\\
(a) & Conv, k = 3, d = 1, p = 1, s = 1 & avg & 95.97 & 81.23 & 43.57 \\
&\\
(b) & Conv, k = 3, d = 1, p = 1, s = 1 & mul & 96.44 & 85.03 & 43.52 \\
&\\
(c) & Conv, k = 3, d = 1, p = 1, s = 1 & avg & 96.10 & 82.44 & 43.11 \\
 & Conv, k = 3, d = 2, p = 2, s = 1 &  &  &  &  \\
 &\\
(d) & Conv, k = 3, d = 1, p = 1, s = 1 & mul & 96.64 & 79.32 & 45.25 \\
 & Conv, k = 3, d = 2, p = 2, s = 1 &  &  &  &  \\
 &\\
 & Conv, k = 3, d = 1, p = 1, s = 1 &  &  &  &  \\
 (e) & Conv, k = 3, d = 2, p = 2, s = 1 & avg & 96.03 & 85.49 & 43.01\\
  & avg pool, k = 3, p = 1, s = 1 &  &  &  &  \\
  &\\
 & Conv, k = 3, d = 1, p = 1, s = 1 & &&&\\
 (f)& Conv, k = 3, d = 2, p = 2, s = 1 &mul & 96.27 & 83.99 & \textbf{46.36} \\
  & avg pool, k = 3, p = 1, s = 1 &  &  &  &  \\
\hline
\end{tabular}
\label{tab2}
\end{center}

\end{table}

\subsection{evaluation with aux loss}
We also use aux loss to assist the training of network. 
In the experimental model listed in B and C part of this section, the aux loss is calculated according to the feature maps after the layer3 block of Densenet. 
Considering that we need to enlarge the feature maps to the same size as the input image to calculate the aux loss, the too small size of feature maps will bring excessive noise. 
In order to further improve the results, we change to calculate the aux loss after layer2 block. 
As shown in Table.\ref{tab3}, calculating after layer2 block is better than after layer3 block, which get 1.8\% improvements on mIoU .
\begin{table}[htbp]
\caption{The results of different aux loss}
\begin{center}
\begin{tabular}{cccc}
\hline
 & \textbf{pacc(\%)} & \textbf{mpacc(\%)} & \textbf{mIoU(\%)} \\
\hline
After layer2 block & 96.27 & 83.99 & 46.36 \\
After layer3 block & 96.75 & 83.83 & \textbf{48.13} \\
\hline
\end{tabular}
\label{tab3}
\end{center}
\end{table}

\subsection{ result comparison}
Through the gradual experiments of the three methods in B, C, D of this section, we achieve 7.7\%, 2.6\%, 1.8\% improvements respectively on major metric mIoU, and get a total of 12\% promotion. 
We compare our model with other models in \cite{wu2018vh} on the PSV dataset. 
The detailed results are shown in Table \ref{tab4}, ours get the advanced result on mIoU, and majority best on the IoU of each class.

\begin{table*}[]
\caption{The segmentation performance of different models on PSV dataset (\%). }
\begin{center}
\begin{tabular}{cccccccccc}
\hline
\textbf{model}& \textbf{background} & \textbf{parking} & \textbf{White solid} & \textbf{White dashed} & \textbf{Yellow solid} &\textbf{Yellow dashed}& \textbf{pacc} & \textbf{mIoU} & \textbf{Model size} \\
\hline
FCN\cite{long2015fully} & 85.88 & 13.16 & 18.42 & 7.40 & 23.09 & 20.32 & 86.18 & 28.04 & 500M \\
FCCN\cite{wu2017fully} & 92.53 & 22.50 & 29.60 & 11.87 & 41.21 & 27.38 & 92.66 & 37.51 & 537M \\
HFCN\cite{yang2018semantic} & 93.87 & 25.46 & 36.26 & 18.97 & 45.08 & 26.87 & 93.97 & 41.09 & 555M \\
VH-HFCN\cite{wu2018vh} & 96.22 & 36.16 & 39.56 & 21.46 & \textbf{47.64} & \textbf{38.03} & 96.25 & 46.51 & 544M \\
ours & \textbf{96.69} & \textbf{38.52} & \textbf{41.43} & \textbf{34.91} & 40.27 & 36.96 & \textbf{96.75} & \textbf{48.13} & \textbf{147M} \\
\hline
\end{tabular}
\label{tab4}
\end{center}

\end{table*}

\begin{figure*}[ht]
\centering
\includegraphics[width=0.9\textwidth]{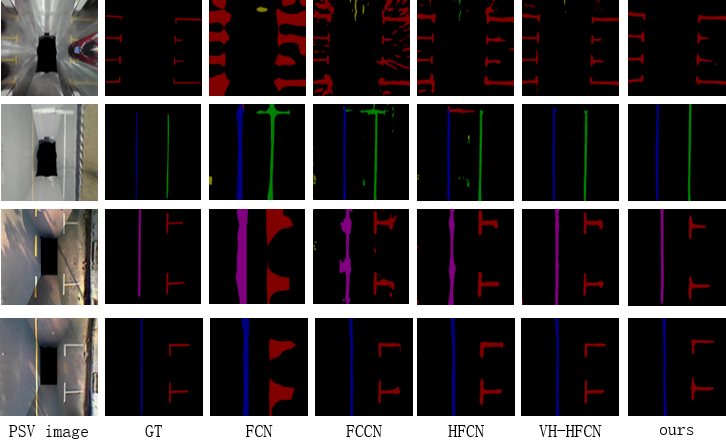}
\caption{Some visual results of segmentation on PSV dataset. The column from left to right represent original image, ground truth, the prediction of FCN, FCCN, HFCN, VH-HFCN, and our model.}
\label{result}
\end{figure*}

In addition, in other models, the accuracy of dashed lines is obviously lower than that of solid lines, and there is quite difference in each class; in our model, there is not much difference. 
What is more, for the model size, ours is 3.7 times smaller than VH-HFCN. 
The smaller size of model can reduce the size of required memory space.
At the same time, smaller size also means less computation parameters, thus it can improve the processing speed of the network.
There are some visual samples of predictions by different model in Fig.\ref{result}. 
We can see that our model can precisely segment the area of lane markings and parking slots, slightly influenced by background noise, and the blank area in dashed line can also be identified correctly.

\section{CONCLUSIONS}
In this paper, aiming at panoramic images of PSV dataset, we segment the lane markings and parking slots by the method of semantic segmentation. 
The two parts of the method we proposed, dynamic loss weights and residual fusion block, are proved to be very effective. 
Moreover, our method is not specially designed for liner structures, which can also be used in general semantic segmentation task. 
Comparing with other models, ours is more precise in result and smaller in model size. 
In future work, we will further improve the accuracy and compress the network model to meet the requirement of using in embedded and mobile platform.

\bibliographystyle{unsrt} 
\bibliography{IEEEabrv,ref} 

\end{document}